\documentclass[final,3p,times,2020]{elsarticle}
\usepackage{amssymb}

\usepackage[colorinlistoftodos,prependcaption,textsize=tiny]{todonotes}
\usepackage{multirow}
\usepackage{rotating}
\usepackage{caption}
\usepackage{graphicx}
\usepackage{hyperref}
\usepackage{natbib}

\hypersetup{
	colorlinks=true,
	linkcolor=blue,
	filecolor=magenta,
	urlcolor=cyan,
}
\title{Digital Twins: State of the Art Theory and Practice, Challenges, and Open Research Questions}

\author{Angira Sharma$^\dag$, Edward Kosasih$^\S$, Jie Zhang$^\S$, Alexandra Brintrup$^\S$, Anisoara Calinescu$^\dag$ \\ $^\dag$\textit{Department of Computer Science, University of Oxford} \\ $^\S$\textit{Institute for Manufacturing, University of Cambridge}}

\vspace{3em}

\begin{document}

\begin{abstract}
Digital Twin was introduced over a decade ago, as an  innovative all-encompassing tool, with perceived benefits including real-time monitoring, simulation and  forecasting.
However, the theoretical framework and practical implementations of  digital twins (DT) are still far from this vision. Although successful implementations exist, sufficient implementation details are not publicly available, therefore it is difficult to assess their effectiveness, draw comparisons and jointly advance the DT methodology.  
This work explores the various DT features and current approaches, and the shortcomings and reasons behind the delay in the implementation and adoption of digital twin.
Advancements in machine learning, Internet of Things and big data have contributed hugely to the improvements in DT with regards to its real-time monitoring and forecasting properties. Despite this progress and individual company-based efforts, certain research gaps exist in the field, which have caused delay in the widespread adoption of this concept. We reviewed relevant works and identified that the major reasons for this delay are the lack of a universal DT reference framework, domain dependence, security concerns of  shared data,  reliance of digital twin on other technologies, and lack  of DT performance metrics.  We define the necessary components of a digital twin required for a universal reference framework; these also validate its uniqueness as a concept, when compared to similar concepts like simulation, autonomous systems, etc. This work further assesses the digital twin applications in different domains and the current state of machine learning and big data within the DT context. It answers relevant questions, and identifies novel research questions, contributing to a better understanding of DTs, and advancing the theory and practice of digital twins. 
\end{abstract}

\maketitle


\section{Introduction}
How can one reduce the cost of producing a prototype and performing tests on it? How can one perform the extreme tests on a prototype which cannot be performed in a laboratory? How can a prototype imbibe all the information and outcome of these tests, to provide an accurate prediction to the future experiments? How can one monitor a physical asset in real-time\footnote{Real-time means at the time frequencies at which the state of a physical asset changes or is expected to change significantly. For example: for an aircraft it could be in seconds, minutes or hours, whereas for a manufacturing unit it could be in hours or days.} and be alerted before anything goes wrong? How can  we  humans have access to this real-time information of all the components involved in a physical asset, perform meaningful real-time analysis on this information, and make timely, robust and efficient decisions for future operations based on them?
The answer -- Digital Twin (DT).

Digital Twins are virtual product models which encompass all the above qualities \citep{Schleich2017}. \citet*{grieves2017digital} define the Digital Twin (DT) as  "a set of virtual information constructs that fully describes a potential or actual physical manufactured product from the micro atomic level to the macro geometrical level. At its optimum, any information that could be obtained from inspecting a physical manufactured product can be obtained from its Digital Twin. Digital Twins are of two types: Digital Twin Prototype (DTP) and Digital Twin Instance (DTI)". DT aims to combine the best of all worlds, namely, twinning, simulation, real-time monitoring and analytics. Digital Twin has been recognised as the next breakthrough in digitisation, and also as the next wave in simulation \citep{Tao2019, Rosen2015}. It can save cost, time and resources for prototyping, as one doesn't need to develop the physical prototype(s), but can perform the same tests on a virtual prototype without affecting the real operation \citep{philipshealth, Roy2016}. 
Gartner, a research and advisory company, listed Digital Twins as one of the "Top 10 Strategic Technology Trends" for 2017, 2018, and 2019 \citep{Cearley2013, Cearley2018, Forni2016}. Furthermore, Market Research Future predicts that the Digital Twin market will reach 35 billion USD by 2025 \citep{marketsandmarkets}.

Despite the stated advantages and potential of DT technology, certain research and implementation gaps exist, which have hindered the adoption and advancement of this concept since its inception around 2003: the diverse applicability of DT is a result of its reliance on advanced evolving technologies, mainly IoT, big data and machine learning.  
The real-time monitoring and data collection capability of a DT is reliant on IoT devices in the environment and enterprise information systems, whereas the analytics is reliant on big data and machine learning tools. Combining these technologies and implementing them for one or more physical assets requires extensive domain knowledge, as would be required while creating any physical asset's physical prototype. 
Furthermore, the DT vision is continuously evolving, as the technology, industry and  customer needs evolve. 
 Not only is the overall concept of DT not fully established, there is no universally accepted definition of DT, and no established standards for implementing it. 
As a technology which is used across various domains and is dependent on other evolving  technologies, it eventually becomes dependent on the current state of these technologies and has to be tailored for each domain as well. Having no standards further  impedes the widespread design, implementation and adoption of this technology.

In this work we attempt to study these gaps, and to propose solutions. As DT is a practical concept, this work emphasises the need for a comprehensive theoretical specification -- via a DT reference framework, leading to an  implementation and evaluation methodology. The contributions of this paper are: \begin{enumerate}

		\item Assess how the dependence on other technologies  is impeding the advancement and adoption of the DT technology and study the involvement of machine learning and big data in it.
	\item Study how domain (or sector or type of industry) affects the DT implementations. 
	\item  Study the gap between the ideal DT concept and practical implementation. 
	\item Evaluate the current state of DT and discuss and answer its limitations and challenges.
	\item Provide a comprehensive analysis of the past efforts at theoretically establishing, and practically implementing  digital twins. 
			\item Deduce the  answer to the debate concerning the definition and implementation of DT.

\end{enumerate}

\textit{Assumptions: }Due to lack of documented implementation methodology of DT, it is infeasible to fully comprehend how the various components such as IoT, data, simulation, real-time synchronisation and machine learning affect the implementation of DT.  Keeping in mind the importance of these components for the concept of DT, in this work we assume good inter-connectivity and compatibility among all these components. 

The rest of the paper is organised as follows: Section \ref{rev} discusses and contrasts the existing reviews. Section \ref{dt} discusses the concept, components, properties, versions, evolution and assumptions of DT. Section \ref{em} discusses the existing theoretical, practical and industrial models of DT across different domains.
It also explores the existing use and potential of machine learning  and data in DT. Section \ref{types} reviews the industries where DT can be particularly useful.   Section \ref{chal} discusses the challenges and limitations with the existing DT models. Section \ref{future} discusses the future directions and poses some new questions for the field, and Section \ref{concl} concludes the paper.

\section{Previous Work} \label{rev}

As DT is a general concept that  can be tailored for a particular domain, there has been a decent amount of literature reviews focusing on  either its implementation in specific domains or in  general. Some works focus on specific domains, such as  manufacturing \citep{Negri2017}, aerospace \citep{Rios2015} and production science \citep{Kritzinger2018}, others are general \citep{Enders, Tao2019}. Most reviews assess the current definition of DT and provide new insights. For example,  \citet{Kritzinger2018}'s categorical review focuses on production science and categorically lists the papers in the areas of Digital Twin, Digital Shadow and Digital Model, whereas \citet{Negri2017}   explores all the papers in industrial manufacturing, mainly differentiating whether big data and data models exist in the literature.  Consequently, the insights of these   reviews are domain-specific, rather than generic and transferable across domains. For example, \citep{Kritzinger2018} looks at similar concepts of Digital Shadow and Model from a manufacturing perspective, whereas \citep{Negri2017} explores data models in a specific class of manufacturing systems.
\citep{Enders} provides a general view on DT. It compares across 87 applications and proposes a classification scheme. However, although such reviews are extensive, they do not include any white papers or industrial works.  As DT is both a theoretical and applied concept, leaving out the practical implementations would provide an incomplete and insufficiently representative description, hence we include these in our review.   \citet*{Tao2019} are among very few authors who explore the specific practical implementations of DT.

The previous works build very little on practical implementations, and focus more on the categorisation of papers and development of theoretical models. The reasoning behind this, as mentioned in the Introduction and further elaborated in  Section \ref{domain}, is that, because DTs are domain-dependent and rely on multiple technologies, the conceptualisation of DT  -- taking the domain into regard and coming up with a sufficiently-detailed and specified, yet universal model -- is a difficult task. 
Nonetheless, these reviews pose the following questions and insights, which help to advance the DT concept further. In this review we discuss these pressing issues in DT:

\begin{itemize}
	\item  The 
	need for a proper definition of DT which could cover the different application domains \citep{lu2020digital, Tao2019}. 
	\item An investigation of concrete case-studies, and a reference model for domain-specific requirements \citep{Kritzinger2018, Negri2017}.
	\item An investigation on how DT research could contribute to the Internet of Things (IoT)  and Information Systems (IS) disciplines 	\citep{Enders}.
	\item The analysis of the debate  initiated in 2013 \citep{Lee2013} and further explored in \citep{Negri2017}, whether DT covers the entire product lifecycle \citep{Rosen2015, gabor2016simulation} or just the product \citep{Schroeder2016, Abramovici2016}.
	\item The scope and need of Big Data and Data Models in DT \citep{Negri2017}, and data-related issues \citep{Rosen2015}.
\end{itemize} 

Apart from the above, in this paper we pose some new research questions and insights, which aim to help exploring and advancing the practical applications of DT.
\subsection{How this review differs}

Previous reviews identified critical  DT insights and questions. This review answers them and poses some new questions, to further advance the DT theory and practice. Also, this review presents the limitations, challenges, and new properties of DT, by researching through the practical implementations of DT, which was previously  under-explored. Compared to the existing reviews, this review does not focus on the categorisation and statistics of the literature based on models, domain, etc., but rather presents the conceptual view of the implementation of  DT concept by looking at the root causes behind the challenges of implementing DT. Also, we address the following novel aspects and questions:
\begin{enumerate}
	\item A universal  reference framework for DT.
	\item The gap between the actual DT concept and implementation (due to no universal definition or architecture).
	\item The importance of real-time machine learning  and data in DT.
	
	\item What is impeding the spread of DT?
	\item How domains affect the implementation of DT?
	\item How  successful the existing implementations of DT are?
\end{enumerate}

In all, this review provides a different viewpoint and analysis to the existing literature and DT implementations. 

\section{Digital Twin} \label{dt}

Digital Twins were first introduced in early 2000's by Michael Grieves in a course presentation for product lifecycle management \citep{Grieves2015}.  In 2011, implementing DT was considered a complex procedure, which required many developments in different technologies \citep{Tuegel2011}.
Despite being coined in 2003, the first description to use of Digital Twin was years later by NASA in Technology Roadmaps \citep{Shafto2010}, where a twin was used to mirror conditions in space and to perform tests for flight preparation (another example of such hardware twin was the Airbus Iron Bird). Dawned with the aerospace industry, soon around 2012 the manufacturing industry  started to use DT. So what took the concept nearly a decade to be implemented?

With the advancements in technologies like cloud computing, IoT and big data, many domains have seen major developments, such as  Industry 4.0 \cite{Lu2017}, Physical Internet \citep{Ballot2014, Sternberg2017}, Cyber-manufacturing \cite{Jeschke2017}, Made in China 2025 \cite{Li2018}, Cloud Manufacturing \cite{Xu2012} etc. Industry 4.0 has seen a revolution mainly because of digital advancements, IoT and Big Data \citep{Negri2017, Jazdi2014}. It was because of the Industry 4.0, the storage of all data in digital format, and sensors being inbuilt into the industrial spaces, that the implementation of digital twin was possible, rejuvenating the concept. Moreover, with the emerging extensive simulation capabilities, it became feasible to perform realistic tests in a virtual environment. Owing to these technical advancements, the idea of implementing a functional DT was soon adopted by companies like IBM, Siemens and GE, as a utility for themselves and for their clients.

\subsection{Defining DT}
On the journey to define the model and properties of DT, various terms have existed  in multiple literature works, such as- 'ultra-high fidelity' \citep{Reifsnider2013} , 'cradle-to-grave' \citep{Tuegel2012}, 'integrated' \citep{Tuegel2012} model , 'integral digital mock-up (IDMU)' \citep{Rios2015}. These terms are important and relevant to the DT concept, however, having multiple definitions and terms has delayed reaching a consensus on a single representative, unifying definition.

In the simplest words, a digital twin is a 'digital' 'twin' of an existing physical entity. What makes a DT all of the above are its properties, which we discuss ahead.

Though the literal meaning of digital twin seems simple at first, the definition of DT has been a subject of debate. For some members of the community (\citet{Schroeder2016, Abramovici2016}) it is correct to consider the Digital Twin as the final product, whereas for others (\citet{Rosen2015, gabor2016simulation})  it is the entire product lifecycle.

Note that, for consistency and generality, we have called the conventionally termed 'product' an 'asset' in the rest of the paper.

\subsection{Components of DT}  \label{components}
DT was first introduced by \citet{Grieves2015} with three components: the digital (virtual part), the real physical product, and the connection between them.  However, other authors, such as \citet{Tao} have extended this concept to have five components, by including data and service as a part of DT.  
\citet{Tao2019} also identify VV\&A (verification, validation and accreditation) as DT components. 
With data models coming into the picture, \citet{Miller2018}  extend the definition of DT to be an integration of multiple models of a model-based enterprise (by creating associations between different models and relations between data stored in different parts, a digital twin can be formed).

As conceptually sound the above definitions are, reaching consensus on a DT definition requires specifying the fundamental requirements for a DT. 
With the advancements in technologies on which DT depends (such as machine learning, big data and cybersecurity) these requirements have changed over time. Moreover, the domain-dependence of DT call for a defining the components which can be generalised across domains.

There have been components and properties which have existed in some works and have been missing in others. We collate this information in the Table \ref{col}. The properties and components which we state are necessary, are a result of their existence in the literature, and our understanding of the DT concept. We thus integrate the contributions of previous works, which have only been concerned with some components of DT, to provide a holistic definition of DT. Based on this analysis and our understanding, we define the elementary and imperative components of a DT.

\subsubsection{Elementary Components}
The elementary components are those without which a DT cannot exist:
\begin{enumerate}
	\item Physical Asset  (could be either a product or a product lifecyle)
	\item Digital Asset (the virtual component)
	\item 2-way synchronized relation between the physical and digital asset (for information flow) 
	
\end{enumerate}

\subsubsection{Imperative Components}
The imperative components add to the properties of DT, to make it the all-encompassing tool of simulation, real-time monitoring and analytics. Without any of these, the uniqueness of DT ceases to exist:
\begin{enumerate}
	\item IoT devices -- to collect real-time information from various sub-components of the physical asset.
	\subitem Requires: High-fidelity  connection between all IoT devices, for accurate and timely flow of information.
	\item Integrated Time continuous Data -- gathered from different IoT components for machine learning and analytics is essential in order to monitor the system, guarantee correct behaviour and provide input to the machine learning system.
	\subitem Requires: Big data analysis and storage tools for extracting useful information from data.
	\item Machine learning--  for predictions and feedback as well as to propose and mitigation strategies, in exceptional circumstances. 
	\subitem  Requires: A joint optimisation feature for all the sub-components of the DT.
	\item Security of the data and information flow among the various physical components of the physical asset.
	\item DT Performance evaluation.
	\subitem Requires: Evaluation metrics (e.g. accuracy, resilience, robustness, costs), and evaluation methods and tests.
\end{enumerate}

\begin{center}
	\captionof{table}{Adjacency matrix showing the various components present and absent in literature ['\textbullet' indicates 
		present and ' ' indicates absent, '*' means indication but not explicitly]}
	\begin{tabular}{ p{1cm}||p{1.15cm}| p{1.5cm}| p{0.5cm}| p{0.8cm}|p{1.6cm}| p{1.5cm}| p{0.7cm}| p{1.15cm}| p{1cm}| p{1cm} }
		
	\small	Papers  & 	\multicolumn{10}{c}{Components} 	 \\ 
		\hline
		\hline
		& \small Transfer of Info. & \small Bijective  relationship & \small IoT & \small Static Data & \small Time-continuous data & \small Statistical Analysis &  \small ML & \small Domain Specific Services  & \small Testing & \small Security    \\
		\hline \hline
		\cite{Shafto2010} & \textbullet&\textbullet& \textbullet & \textbullet & \textbullet & \textbullet & \textbullet & \textbullet & \textbullet *& \textbullet\\ \hline 
		
		\cite{Tuegel2011} & \textbullet& \textbullet& \textbullet& \textbullet& \textbullet* & \textbullet& \textbullet * & \textbullet& \textbullet * &   \\ \hline
		
		\cite{Glaessgen} &\textbullet& \textbullet& \textbullet & \textbullet & \textbullet & \textbullet & \textbullet & \textbullet & \textbullet * &   \\ \hline
		
		\cite{Rios2015} & \textbullet&  & \textbullet& \textbullet&  & \textbullet&  & \textbullet&  &  \\ \hline
		\cite{Rosen2015} & \textbullet& & \textbullet* & \textbullet& &  & & \textbullet& &  \\ \hline

		\cite{Grieves2015} & \textbullet& \textbullet& \textbullet& \textbullet& \textbullet& \textbullet& \textbullet& \textbullet&   &   \\ \hline  
		
		\cite{Schroeder2016} & \textbullet&   & \textbullet& \textbullet&   &\textbullet&   & \textbullet&  &  \\ \hline
		
		\cite{Schleich2017} &\textbullet&  & & & & \textbullet& && \textbullet& \\
		\hline
		\cite{Negri2017} & \textbullet& \textbullet& \textbullet& \textbullet&  & \textbullet& & \textbullet& &  \\
		\hline
		
		\cite{Kritzinger2018} & \textbullet& \textbullet& \textbullet& \textbullet &  & \textbullet& & & & \\ \hline

		\cite{Tao} & \textbullet& \textbullet & \textbullet & \textbullet & \textbullet& \textbullet & \textbullet & \textbullet&   &   \\ \hline

		\cite{Droder2018} &   &   & \textbullet&   &   & \textbullet&  &    & \textbullet* &  \\ \hline
		
		\cite{Liu2018a} & \textbullet & \textbullet& \textbullet& \textbullet& \textbullet * & \textbullet& \textbullet * & \textbullet&  \textbullet * &    \\ \hline

		\cite{Tao2019} & \textbullet& \textbullet& \textbullet& \textbullet*&  & \textbullet&  & \textbullet& \textbullet& \textbullet\\
		\hline
		
		\cite{Enders} & \textbullet& \textbullet& \textbullet * &  * &   &  * &   & \textbullet&   &   \\ \hline  
		\cite{Min2019} &\textbullet& \textbullet& \textbullet & \textbullet & \textbullet & \textbullet & \textbullet & \textbullet & \textbullet * & \textbullet\\ \hline
		
		\cite{Cronrath2019} & \textbullet& \textbullet& \textbullet& \textbullet& \textbullet*& \textbullet& \textbullet&   & \textbullet& \textbullet* \\ \hline
		
		\cite{qi2019enabling} & \textbullet& \textbullet& \textbullet& \textbullet& \textbullet*& \textbullet& \textbullet&  \textbullet & \textbullet*& \textbullet* \\ \hline
		
		\cite{grieves2019virtually} & \textbullet & \textbullet & \textbullet & \textbullet & \textbullet* & \textbullet&  & \textbullet* &  \textbullet* & \textbullet \\ \hline
		
		\cite{lu2020digital} & \textbullet & \textbullet & \textbullet & \textbullet & \textbullet* & \textbullet &  & & & \textbullet
		\\

	\end{tabular}
	
	\label{col}
	
\end{center}

\begin{center}
	\captionof{table}{The different components of DT and their key roles}
	\begin{tabular}{ l|l }
		Component & Role \\
		\hline
		Physical Asset & what the digital twin is a twin of\\
		Digital Asset & the digital twin \\
		Continuous Bijective Relation & for real-time synchronisation and twinning \\
		IoT & for data collection and information sharing\\
		Time continuous data & for synchronisation and input to machine learning \\
		Machine learning & for analytics of the asset\\
		Security & to prevent data leaks and information compromises\\
		Evaluation metrics / Testing & to evaluate the performance of DT 
	\end{tabular}
	
	\label{comp_table}
	
\end{center}
\noindent Table \ref{comp_table} summarises how each component contributes uniquely to the functions of DT. Removing any component voids the DT of the functionality and its uniqueness, as discussed ahead.

\subsection{Properties of a DT} 
As simple as the definition sounds, the properties of the DT are what makes it more than just a 'digital' 'twin'.
\subsubsection{Necessary Properties}
These are the properties inherent to any digital twin:
\begin{enumerate}
	\item Real-time connection with the physical entity (\cite{Rios2015} defines a 'biunivocal' relation between DT and the physical asset). 
	
	\item Self-evolution: A very important property which has not been explored much (introduced in \citep{Tao}). With this DT can learn and adapt in real-time, by providing feedback to both physical asset (via the human asset) and DT. This can be easily harnessed now due to the uprise of machine learning tools: to remodel and redesign itself (such as reinforcement learning). The frequency of this synchronisation depends on the update scenarios, such as event-based (supply chain), periodic intervals (aircraft), condition based (logistics), etc.

	\item Continuous machine learning analysis (dependent on the frequency of the synchronisation), not just one-time output forecasting.

	\item Availability of time-series (or time continuous) data for monitoring  \citep{Tao}, and as input to machine learning system.

	\item Domain dependence (or Domain specific services): According to the domain, a DT may provide or prioritise services specific to the industry. These are the same 'domain specific' services which exist in the physical asset (for example the optimisation problems for an aircraft and a manufacturing unit will prioritise or add more weight to different parameters).

\end{enumerate}

\subsubsection{Dynamic Properties}
Based on these dynamic properties, a hierarchy of digital twin can be created: 
\begin{itemize}
	\item Autonomy: A DT (or for that matter any information provider \cite{Cortada2018}) could either make changes to the physical asset itself, or a human in control could make changes to the DT. This applies differently to  different hierarchies of components present in the twin, such as to some parts of the machine learning system, or some part of the decision making system. Hence, the property of a DT to be autonomous, not autonomous, or partly autonomous is case-dependent.This classification also includes the self-evolution mechanism of DT (what changes must it make to itself, and what changes must be approved by a human).
	\item Synchronisation: Synchronisation of data could either continuously or at certain time intervals. These depend on a number of factors such as technology, resources available, need for the data and  type of machine learning algorithm being used. A DT could have sub-components which could be partly continuously synchronised and partly event-based synchronised.   
	\subitem This synchronisation can result in different hierarchies based on the following:
	\subsubitem -How often the data is collected?
	\subsubitem -How often the data is stored?
	\subsubitem -How often the DT is updated? (note that this is different to the property of autonomy as that deals with 'who' rather than 'when')
\end{itemize}

\begin{figure}
\centering
	\includegraphics[scale=0.4]{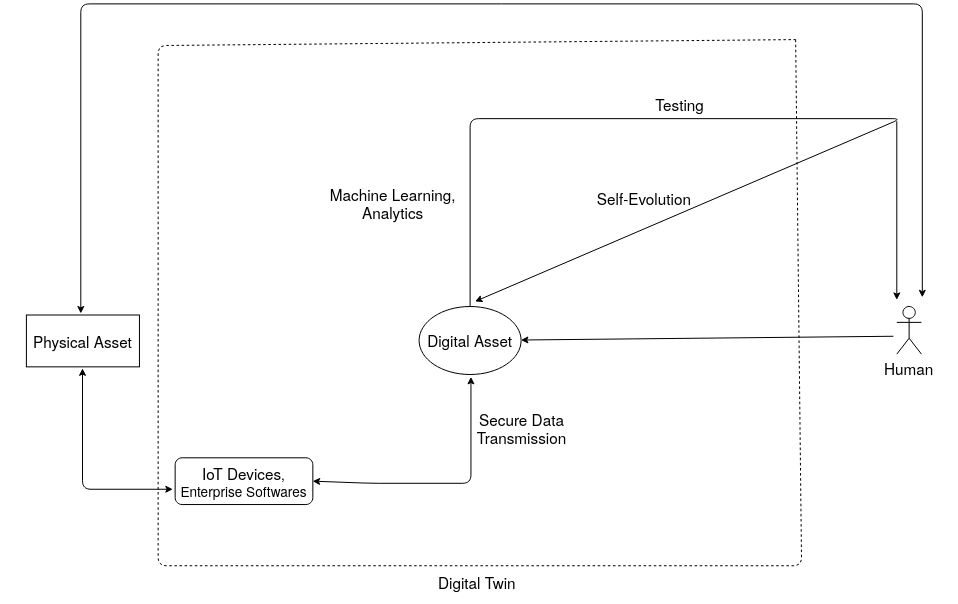}
	\caption{DT reference framework with components and information flows:\\ \textbf{Flow of Information} - IoT, Data and Machine Learning
		Data will be gathered from various IoT devices and sent to the DT. DT will use this data for machine learning and analytics. The prediction from DT will be provided to the human, to take further steps. Details of these further steps could be fed back into DT, to test how their corresponding actions will affect the DT.	}
	\label{diag}
\end{figure}

\noindent Based on the information present in current literature and our analysis, we present a reference framework in Figure \ref{diag}.

\noindent Answering the debate concerning whether a DT should be a product or product lifecycle, after analysing the literature our take on it is that, a DT can be both, for example, DT of a car or DT of the car's production lifecycle. This stems from the components of DT defined in Section \ref{components}; as long as all the components are present, any physical asset could have its Digital Twin.

\subsection{How is DT different from existing technologies}

The diverse applications of DT such as simulation, real-time monitoring, testing, analytics, prototyping, end-to-end visibility \citep{Ivanova}, can be broadly classified as sub-systems of DT (for example, a DT can be used for testing during prototyping, for real-time monitoring and evaluation, or for both). It is the presence of all the components discussed in the previous section that makes a  DT different from these, as described in Table \ref{dif}.

\begin{center}
	\captionof{table}{How DT differs from existing technologies}
	\begin{tabular}{l|l}
		Technology & How it differs from DT \\
		\hline
		Simulation & No real-time twinning  \\
		Machine Learning & No twinning \\
		Digital Prototype & No IoT components necessarily \\
		Optimisation & No simulation and real-time tests \\
		Autonomous Systems & No self-evolution necessarily \\ 
		Agent-based modelling & No real-time twinning \\
	\end{tabular}
	\label{dif}
\end{center}

\subsection{A brief overview of other similar concepts that preceded DT}

The concept of digitising and twinning are not new. Many similar concepts have preceded DT, however, for the reasons described briefly below they differ. 
\begin{itemize}
	\item \textbf{Digital Shadow, Digital Model} - \citep{Kritzinger2018}  A Digital Model has only manual exchange of data and it does not showcase the real time state of the model. Digital Shadow is a saved data copy of the physical state, with one-way data flow from physical object to the digital object. The digital twin on the other hand, has fully integrated data flow where the digital twin properly reflects the actual state of the physical object.  
	\item Semantic \textbf{Virtual Factory Data Model} (VFDM) are virtual representations of factory entities \citep{Terkaj2014}. These were used in manufacturing and industrial spaces \citep{Terkaj2012virtual}. DT differs from VFDMs because of the property of real-time synchronization. On one hand, where VF is a data model only, on the other, DT is real-time and synchronised. 
	\item 	\textbf{Product Avatar} is a distributed and decentralised approach for product information management with no concept of feedbacking and may represents information of only parts of the product \citep{Rios2015}
	\item \textbf{Digital Product Memory}: \cite{Miller2018} see DT as an extension of semantic / digital product memory where a digital product memory  senses and captures information related only to a specific physical part and thus can be viewed as an instantiation of DT. 
	\item  \textbf{Intelligent Product} \cite{Mcfarlane, Wong2002, Meyer2008}: A DT can be seen as an extension of Intelligent Product which uses the new technologies like IoT, Big Data and Machine Learning which were lacking in the concept of Intelligent Product.
	\item \textbf{Holons} \cite{McFarlane2000, Valckenaers2007, Valckenaers2009}: As an initial computer-integrated manufacturing tool, holons formed the basis for all the technologies described above. 
\end{itemize}

\noindent Clearly DT has many novel and exclusive collection of applications to provide. Having established its background and components, we now process to its models in the next sections.

\section{Existing Models} \label{em}
\subsection{Theoretical models} \label{theo}
In this section we explore the various attempts  to define the architecture of DT. Most of these are for particular domains/sectors, while few are generalised.

The most popular sector for implementing a DT is  Product Lifecyle Management (PLM). This is chiefly because  DT provides a holistic view to the widespread components of PLM which is useful to solve the existing problems in this domain \citep{Kiritsis2011}; although challenges persist, such as  \citet{Tao} discuss how data in PLM is isolated, fragmented and stagnant which poses issues. The work also  presents a theoretical framework of applying DT to PLM by proposing three design steps: conceptual design, detailed design and virtual verification.  The work gives an example of DT for bicycle as a case study, but does not mention the implementation methodology. Such as  to perform tests on the bike with parameters involving  brakes, speed, customer weight, etc., proper physics simulators are needed and these simulation softwares or frameworks have to be defined. 

\citet{Liu2018a} propose using Unity3d open source engine for implementing DT. They create a reference model to handle and synchronize complex Automated Flow-Shop Manufacturing System (AFMS) systems using DT and deals with decoupling multi-objective optimization problems. As mentioned in Section \ref{dt}, if DT is an entire product lifecycle, then analytics and using machine learning in DT would mean defining a  universal joint optimization problem for the lifecyle. Handling multitude of parameters, and solving multi-objective optimisation problems is an existing challenge for complex systems. 	 However, \citet{Liu2018a} apply the sheet processing DT prototype successfully to Chengdu, China, and use heuristic measures such as production packages, unified system cost, unified system performance to evaluate the performance of DT (resulting in a domain-dependent DT).

This domain dependence property is also discussed in  \citep{Enders}, where a generalised architecture of DT is presented by defining six dimensions for defining the concept of DT -- industrial sector, purpose, physical reference object, completeness, creation time, and connection (some of which were inspired from previous works).  The other  important self-evolving property of DT is explored in \citep{Tao}, where the authors discuss many cases  of using DT in product design and manufacturing.

Continuing the debate on whether DT should be the product or the product lifecycle, \citet{Schleich2017} propose a DT reference model for production systems, for geometrical product specification (GPS), where they pitch the idea to represent the DT as an abstract form of the physical product, mentioning clearly that DT is the entire product lifecycle and not just the product.

As Digital Twins can be considered as artificially intelligent systems, they are closely linked to autonomous systems. This link is explored in  \citep{Rosen2015}, where the authors discuss the importance of DT in simulating the autonomous system's decisions. As autonomous systems need real-time information, the paper argues that digital twin will be very useful and efficient for the autonomous systems to function properly, as DT can collect all data and prior knowledge needed. They discuss the driving aspects for the future of manufacturing, which are: Digital Twin, modularity, connectivity and autonomy. These aspects are also important for the concept of DT -- to see the individual components of DT and then integrate them into a single model. The authors also support the claim that DT is the next wave in simulation,  as it encompasses real time operation and data.  Despite of a well-explored evaluation, the examples considered in the \citep{Rosen2015} paper for implementing a real DT are limited to using DT as a memory buffer to draw data, and do not explore the machine learning and analytics potential. 

Despite being conceptually sound and well-explored, the current DT literature does not fully address the gap between theory and practice, due to the challenges which we discuss ahead in Section \ref{chal}.

\subsection{Practical models and their domain-dependence} \label{domain} \label{prac}

The applications of DT spans across various domains from manufacturing, aerospace to cyber-physical systems, prognostics, health and management, etc. The description of these practical models in literature is scanty. Moreover, eventually, as hard and complex implementing a DT for large systems becomes, these descriptions become vague. Not to forget, this complete methodology is domain dependent as well. We discuss these practical attempts and the affect of domain on these attempts ahead.

Owing to the number of parameters and intricacies of the operations involved 
the most complex but most beneficial implementation of a DT is in the aerospace domain.  \cite{Glaessgen} discusses application of DT for NASA and U.S. Air Force vehicles. The main advantage of using a DT for aerospace is that a DT can  replicate the extreme conditions (thermal, mechanical and acoustic loadings) which cannot be physically performed in a laboratory because laboratory tests cannot go below or above a certain limit. Moreover, conventional approaches in this domain do not consider the specific material involved and type of the component during testing, which is highly desirable. On the other hand, a DT can be  tailored according to tail number to take into account the specification of the materials and its types.
Additionally, the current analytics approaches are mostly based on statistical distributions, material properties, heuristic design properties, physical testing and assumed similarities between testing and real operation. But, as the authors say, these techniques are unable to represent the future extreme requirements, whereas the DT can use maintenance history and other historical fleet data to continuously forecast health and probability of mission success.  A DT can also test the future extreme requirements in real time, as and when they appear. This real-time support is needed as  external support is not always possible. Also, the machine learning capabilities of a DT can make predictions and recommendations for in-flight changes to a mission. A DT also provides self-mitigating mechanisms or recommendations. It is because of the plethora of advantages of using a DT, mainly as an alternative form of testing which can't be performed in the lab, that DT in aerospace is highly desirable, nonetheless, hard to implement because of the multitude of parameters involved.

\citet{Rios2015}  presents a review of using DT in aerospace where the authors describe that, because a commercial aircraft may have more than half a million different component references, it is challenging to create a bijective relation between a particular physical aircraft and its unique digital twin. 
The complexity in implementation of DT also arises due to the interoperability issues among the different softwares used in production, such as PLM, Enterprise Resource Planning (ERP) , Manufacturing execution system (MES), Computer-aided technologies (CAx), etc.  
The work suggests to make use of each product's individuality such as Manufacturers Serial Number (MSN), the EPC (electronic product code), Tail Number (TN), aircraft registration, VIN (vehicle identification number) to create an effective DT, which is similarly difficult because of the number of parameters involved. 

On the other hand, \citet{Tao2019} discuss how prognostics and health management (PHM) have the biggest advantage of using DT as the concept considers ultrafidelity, behaviour and rules modelling, and merges physical, virtual, historical and real-time data to provide trends, optimization and maintenance strategy.  The work also discusses the challenges pertaining to the cyber-physical fusion which occurs while implementing DT; these challenges are: security, robustness, applicability, data acquisition, mining and collaborative control.

The advantages and challenges of using a DT are thus different for different domains. Therefore, a DT has to be tailored according to these different sectors. These differences are explored in \citep{Negri2017}. This work discusses the history of DT and highlights briefly the expectations from DT in different areas such as CPS (focus is on avoiding failures, support health analysis of systems and deformation of materials in the physical twin,  and study the long term-behaviour in different environments), aerospace (maintenance and intervention needs of the aircraft with the use of Finite Element Methods (FEM), Computational Fluid Dynamics (CFD), Monte Carlo and Computer-Aided Engineering (CAE) applications-based simulations), manufacturing (to simulate complex and numerous parameters of the system) and  robotics (to optimise control algorithms during development phase). These highlight how a DT is expected to behave differently as best suited for the different domain.

To the best of our knowledge, the use of ML techniques in DT is next to none. Only \cite{Min2019} lays down a proof of concept for implementing DT with machine learning, in the petrochemical industry. The paper provides some time series data preprocessing solutions for unifying frequency of  time series  data, resolving time lag issues between time series data, reducing data dimensions, regenerating new time series data, etc. They have provided few implementation details: the authors use many production service systems for data gathering and training the digital twin (such as supervisory control and data acquisition (SCADA), programmable logic controller system (PLC), manufacturing execution system (MES), laboratory information management system (LIMS), distributed control system (DCS)), they access IoT data through APIs, and use simulation and optimization systems namely advanced process control (APC) and real-time optimization (RTO).  The DT is not tested by defining metrics but the prediction model is tested using
four evaluation criteria: the model accuracy ratio (MAR), the root mean square error (RMSE), the variance interpretation rate (VIR), and Pearson’s correlation coefficient (PCC).

Since DT is a twin of the physical asset, implementing it requires the same amount of knowledge, that is required when creating the physical twin. For any domain, it is essential to have the domain knowledge to understand the intricacies involved in building and operating the physical asset, such as, how different components link together, how much to weigh parameters during optimisation, etc. Hence, domain experts are  essential to implement a DT.  \cite{philipshealth} suggests the use of adaptive intelligence for DT which demands a great deal of human expertise, as human knowledge forms the cardinal element for this artificially intelligent system,.

The above analysis of DT in  various sectors indicates how DT is domain-dependent and so are the related challenges which we discuss ahead.

\subsubsection{Similarity \& Differences Across Different Domains}
To further lay a clear foundation of the concept of DT, we discuss a the similarity and differences that DTs possess when being implemented across different domains: \\

\noindent \textbf{Similarity} The components of DT mentioned in Section \ref{components} should remain the same (though the level of implementation is subject to domain). \\ \\
\textbf{Differences} The following domain dependent questions and implications arise:
\begin{enumerate}
	\item  Particular domain poses greater challenges for the implementation of DT:
	\subitem Some digital twins might be more feasible and easier to implement than others (due to scale, resources required and number of components). Same goes for data collection. For instance, the aerospace sector has numerous components to handle or a supply chain on a global level has many parameters and multiple inventory points; these cases will prove to be a challenging joint optimisation problem. 
	\item How realistic the  implementation of the DT is with respect to the actual physical asset:
	\subitem Depending on the complexity of the physical asset, the implementation of the DT will differ from domain to domain. The gap between the ideal design and practical design might be large for domains like aerospace and supply chain (w.r.t simulation this is  discussed in  \cite{Vieira2020}), and small for PLM and health and prognostics.
	\item Evaluation of the implementation of DT:
	\subitem The evaluation metrics for performance of DT are dependent on the domain. As some domains will lay more importance on some sub-components such as either of prediction or real-time monitoring, consequently, the performance evaluation of the DT will be subject to the parameters most important for the specific domain (such as for aerospace it is mitigation plan).
	
	\item The interoperability issues in the range of different softwares being used:
	\subitem If  domains use particular type of software (SCADA, ERP, etc) for sub-tasks of product and logistics management, how to make the DT software compatible with these systems, which are specific to the  domain sectors.
	\item The effect of domain knowledge on the implementation of DT:
	\subitem Domain experts are imperative and essential for designing a DT. A thorough knowledge of the particular domain and its every sub-component is crucial for designing a DT.
	\item What is stopping the spread implementation of DT across a domain:
	\subitem Is it the lack of liaison between domain experts and DT providers?
	\subitem Are some domains too complex to implement DT or need a simpler model?
	\subitem Are the lack of compatibility within the domain specific components concerning IoT, data and machine Learning a problem?
	\subitem Is the regulation and standards across domains an issue?
\end{enumerate}

\subsection{Industrial Implementations} \label{indust}
There are a number of companies now which are investing in DT technology, or providing DT software for clients, or using DT functionality for themselves:

\begin{enumerate}
	\item Investing in DT
	\begin{itemize}
		\item Signify Philips is exploring the concept of digital twin for lighting \citep{philipslight} by digitising lighting. They claim it to offer emergency services, real-time monitoring and predictive maintenance. 
	\end{itemize}
	\item Providing DT as a service
	\begin{itemize}
		\item Philips is also providing DT technology for the use of healthcare systems \cite{philipshealth} to get early signs of warning regarding technical issues in Magnetic resonance imaging (MRI), computerized tomography  (CT) scan like medical systems. This could save the downtime that faulty technical systems have on clinical spaces.
		\item IBM is transforming the Port of Rotterdam using Digital Twin \cite{ibm_port} for monitoring and efficiency. IBM is also providing a DT software for PLM \cite{ibm_exch}. Along similar lines, Siemens has a model to implement the digital twin of  power grid in Finland \cite{SiemensAG2017}, \cite{sie_fin} and another for Red Bull in Formula 1 racing \cite{f1}.
		\item Companies, namely, Dassault System's  3DEXPERIENCE \cite{dassault} (which has also built digital twin of Singapore \cite{sing}), AnyLogic \cite{anylogic}, Ansys \cite{ansyss}, Visualiz \cite{visualiz}, PwC \cite{pwc}, Bosch \cite{bosch}, SAP \cite{sap} and Azure \cite{azprod}, \cite{azdt} provide DT implementation software for clients  (though the use of ML in these softwares isn't clear). Apart from these, Oracle provides DT simulator as part of its cloud service \citep{Oracle2017, Corporation}. GE has contributed to the DT literature and holds 2 patents for DT \citep{DellAnno, ge_p} and has a  commercial software Predix \citep{gedt}. The open source community has also explored the concept of DT and, an open source technology Eclipse Ditto implements IoT based DT models \citep{ditto}.
	\end{itemize}
	
	\item Using DT for own use
	\begin{itemize}
		\item DHL has implemented first digital twin supply chain for Tetra Pak's warehouse in Asia Pacific in Singapore \cite{dhl_tetra}.  Though this is real-time monitoring, the use of machine learning seems unclear and is plausibly next to none. 
		\item BP has employed a surveillance and simulation system called APEX for creating virtual copies of its production systems (again ML use unknown) \cite{BritishPetroleum2018}. \\
		
	\end{itemize}
\end{enumerate}

\noindent To the best of our knowledge, none of the above softwares are implemented at global level (but only at country or city level), and more detailed and concrete analysis on this is subject to further investigation. 

\subsection{The gap between the ideal DT and practical DT}
Depending on the technologies such as IoT, big data and machine learning, there may be a huge gap between the ideal implementation of DT and the practical one (such as whether the required advancement in technology is currently available or is subject to further research). Cost and number of available resources could also contribute to increase in this implementation  gap. 
This discussion is subject to more investigation and availability of the actual DT software and its initial design model.

From the previous sections, we do not have any information on the level of implementation. Some use outsourced DT software, but do not describe the frameworks.
 It is hard to know without evaluation metric-based evidence how successful the current implementations of DT are. However, most papers have claimed the following to be successful uses of DT --- data handling, real-time monitoring, simulation testing and optimisation \cite{Tao2019, Min2019}. However, as mentioned, these implementations without all the components described in Section \ref{components} do not fully adhere to the  conceptual definition of DT. Therefore, it is necessary to have a consensus reached regarding a unified, standard architecture of DT.


\subsection{Machine Learning in DT} \label{ml}

Conventional knowledge-based methods are based on one-time machine learning output, whereas a digital twin is a continuous interactive process.  Real-time machine learning capacity is what differentiates a DT from a simulator or a real-time monitoring tool.  Analytics is essential, as one of the cardinal uses of a DT is to be able to reliably and accurately output how a physical asset would behave in conditions which have not arisen yet, using the real-time data it is receiving -- in other words --  for 'testing' the physical asset in an unforeseen situation. The other cardinal use of ML in DT is to predict an impending problem which needs attention.

One work that explores the digital twin with machine learning is  \citep{Min2019}. It presents a proof-of-concept for using machine learning with digital twin in the petrochemical industry, by making use of ML algorithms such as random forest, AdaBoost, XGBoost, gradient boosting decision tree (GBDT), LightGBM, and neural networks. However, information on the exact implementation methodology and software for DT, and how the entire proof-of-concept was implemented in real time feedback loop, is absent. 

Another work, \citep{Droder2018}, discusses using a simulation software along with machine learning for a product, and terms this as 'the digital twin approach'. This is the result of  inconsistency in a unified architecture and definition of digital twin as calling a simulation software with no real-time synchronous  connection to the physical product as a digital twin might be correct according to some definitions while wrong according to other. Hence, it is important for a technology to have all the components deduced in Section \ref{components}, to call it a DT and differentiate it from other technologies.

The 'self-evolving' nature of DT, where a DT can improve itself from the true results of its predictions, can only be implemented using machine learning. Machine learning can also help in creating resilient DT. The promising work \citep{Cronrath2019} uses machine learning in DT not as a feedback mechanism but as an error resolver rather than an analytical tool.  \citet{Cronrath2019} use reinforcement learning to make the digital twin resilient to data or model errors, and to learn to fix such inconsistencies itself.

Despite still being popularly marketed as a  DT software by companies like IBM \cite{ibm_exch}, SAP \cite{sap}, Siemens \cite{SiemensAG2017}, the published literature on using machine learning for Digital Twin is scanty and the extent of use of ML by these companies is uncertain.  

There are many advantages  of implementing an ML model, however, the following technical challenges are imminent to these models, which we suspect might be the reason for lack of ML in the existing DT models:
\begin{enumerate}
	
	\item Real-time synchronisation for continuous feedback:
	\subitem High-fidelity synchronisation between components is required for real-time feedback which depends on the IoT devices and network connection. 
	
	\item Joint optimisation issues:
	\subitem As described earlier, multiple parameters in the network present a complex multi-objective optimisation problem. For very large firms, nailing down the sub-optimisation problems to a single one, can be a challenging task. This requires knowledge of both the domain and  machine learning.
	\item Challenges related to data required for machine learning (discussed in next section)
	
\end{enumerate}

\subsection{Big Data in DT} \label{data}

DT is used in sectors which have multiple components resulting in multiple parameters. Hence the data collected from these sources ends up being a large high-dimensional dataset. Moreover, if the time frequencies of the data collected from these different components do not match, the resulting  data can be fragmented. Therefore, there exist time lags in time series data. Additionally, collection of data from multiple inter-connected and not-connected components, with high-stream synchronisation  and integrating this data, is a challenging task in terms of technology, implementation, cost and resources \citep{Min2019}. 

\citet{Min2019} identify two major issues and sub-issues related to implementing digital twin in real world with its data component:
\begin{enumerate}
	\item As a dynamic environment requires a well-researched tool, better concrete and practical frameworks are needed for big data application to the continuously changing environment of DT. 
	\item Data processing issues for time series data: Data gathered from IoT devices in the  factory  have large dimensions. Moreover, the data collected may have  different time cycles.

\end{enumerate}

The authors identify the above problems and solve the problem of using data from different time frequencies by proposing a method to generate same frequency time series data. 

\citet{Tao2019} make an attempt to define the steps for  data preprocessing, data mining and data optimisation for DT, which are essential for large datasets.  Another work, \citep{Tao}, discusses the limitations of gathering large datasets. The work also assesses the problems  in data management specific to the PLM domain, such as   existence of duplicate data, absence of big data analysis and existence of disintegrated data in different phases of PLM. The paper proposes DT as a solution for comparing this inconsistent data with real values in PLM. 
\citet{lu2020digital} points that domain knowledge could help dealing with missing data issues. \\

\noindent Despite being domain-dependent, there are issues regarding data in DT which exist across domains:
\begin{enumerate}
	
	\item Gathering data from different IoT devices: Depends on the IoT technology, network connection and interconnectivity among different components. 
	
	\item Handling large datasets: Requires proper resources for storage and preprocessing steps.
	\item High dimension of the datasets: Requires expertise in big data and data reduction techniques.	
	\item Synchronising data from different components frequency-wise: Requires standardised IoT devices, connection and research.
\end{enumerate}

\citet{qi2018digital} compare and contrast Big Data and DT, within the context of smart manufacturing \& Industry 4.0. They consider that the main role  of DT is the cyber-physical integration, so that DT maintains an accurate representation of the real system, which   could then be used to predict and feedback optimisation decisions to the real system.

\citet{Huang2020} present a blockchain solution to deal with the problem of data integrity in a system. Despite being promising for health systems, financial systems and cross-industry collaborations, this concept adds to the complexity of data handling. Adding blockchain technology to DT and optimising it with the rest of the components would be the next challenge \cite{Lu2019a} after integrating big data solutions in DT.

\section{Industries where DT can be majorly beneficial} \label{types}
Though DT is a technology which benefits any industry in general, there are certain sectors which can benefit majorly:
\begin{enumerate}
	
	\item Industries where  creating physical prototypes is expensive, requires resources and is time-consuming (such as aerospace, supply chain, manufacturing): Rather than spending time and money for building multiple prototypes for testing a product, digital twin offers a much more efficient and feasible solution.
	
	\item Industries in which extreme testing is required and performing such tests is hard/not possible in the labs (such aerospace, PHM): Tests which cannot be performed in the lab can be simulated by the DT.
	
	\item Industries which require real-time monitoring and mitigation plans for dealing with 'emergent behaviour' \citep{grieves2017digital} (such as health systems, supply chain): Keeping an eye on the real-time status of the physical asset, and being alerted through predictions for an imminent problem can be both efficient and effective. This is especially useful for those organisations which need to make very quick decisions to prevent huge losses.  
	\item Industries with multiple parameters, which could be optimised jointly (such as manufacturing, supply chain): For very large organisations, maintaining and monitoring all sub-components can be an extremely difficult task. Real-time monitoring of all sub-components and joint holistic analytics on such huge models can be beneficial (this goes along the obstacle of 'siloing' presented in \citep{grieves2017digital}, where due to multiple sub-domains in the system, information remains fragmented).
\end{enumerate}

 Despite being dependent on multiple technologies (explored in detail in \citep{qi2019enabling}) which requires experts and resources, DT can lead to huge cost reductions for the one time investment \citep{grieves2017digital}.
DT can enable shorter design cycles  \citep{GeryCTO}, save cost, resources and time on prototyping \citep{philipshealth, Roy2016}, and predict impending dangers in time to mitigate them. This cost reduction could possibly be used as a metric of the performance of DT for profit-oriented companies, i.e., the DT-enabled cost reductions.

\section{Current Challenges and Limitations in DT} \label{chal}

\subsection{Challenges }
Current models face the following challenges, some of which are weighed more depending on the domain the DT is being implemented. These challenges are mainly technical:
\begin{enumerate}
	
	\item Data handling (high dimensional \citep{Schleich2017}; time series, multi-modal and multi-source data communication \citep{Tao}: In large organisations the time-series data is collected from numerous IoT devices, resulting in high dimensions for the dataset. Collecting data from a considerably large number of IoT devices, collating it according to time frequencies and preprocessing it for input to machine learning  is a challenging big data task.
	
	\item High-fidelity 2-way synchronisation is especially hard for large-scale industries, requires resources and high-stream IoT connection \citep{Grieves2014, Schleich2017, Tao}.
	\item Simulation  software and simulation-based optimisation: Building a software for DT  demands a team of programmers, developers and domain experts to test the suitability of the software for the particular task. Moreover, simulation-based optimisation provides faster and efficient solutions \citep{Alrabghi2015}.  Conventionally, solutions calculated by analytical models are fed to the simulation software manually rather than optimisation being done on the simulation software by use of mechanisms like reinforcement learning. 
	
	\item Joint optimisation for all the sub-problems: Depending on the size of the industry, the number of parameters involved in the machine learning optimization problem can be huge, as each industry has smaller optimization problems which are solved at various sub-levels. Combining these sub-problems  into one optimization problem and choosing the right machine learning tools is a challenging task. True predictions of the entire physical asset can only be made if all the components of the asset are taken into consideration, resulting in a joint machine learning-based optimisation problem. This kind of unified ML model will easily facilitate and weigh priorities in a system, and hence perform better analytics.

		\item Global Implementations  (currently only city level or country level implementations exists): Current implementations (as discussed in section \ref{indust}) are mainly focused on city/country. For larger systems, such as supply chain networks and logistics, etc, global implementation level is more desirable. 
	\item Deep learning with DT: Implementing deep learning solutions requires computational resources, expertise and research \citep{Liu2018a}. Managing high dimensional data, with the various other software used by an industry and combining these with  expert deep learning skills and equipment is a tedious task.  New methods like 
continual learning \cite{Lopez-Paz} and federated learning \cite{Yang2019} seem promising for the use of DT but require further research.

	\item Interoperability with existing softwares being used in a production lifecyle \citep{grieves2019virtually}: Industries use multiple software for inventory, product management, etc. The compatibility of DT with these is a challenging issue, tackling which might lead to delay in implementations.
	
	\item Cybersecurity concerns, IoT security, cross industrial partners security \citep{grieves2019virtually}: With the digital twin operating across multiple industrial partners and inventory sites, the security concerns are inevitable. Not only the cross industry security concerns but also the leak of real-time monitoring data can be hazardous to a firm.

\end{enumerate}

\subsection{Limitations}
Despite the advantages that DT brings to an industry, depending on various factors, there exist certain limitations:

\begin{enumerate}
	\item Cost: Since implementing DT and profiting from it is a timely process, DT can be costly if the life and span of a project is small.
	\item Complexity: As DT requires interoperability among various components, real-time tools, formulating a joint optimisation problem and big data resources, putting these together can be time-consuming for an industry and may lead to unwanted distractions. 
	\item Research Add-on: Like any technology, DT will be needed to be updated according to the recent developments in the technologies it relies on (IoT, big data, machine learning). Industries with long-term DT use will therefore need to continuously invest in this research, which might lead to added cost.
\end{enumerate}
\section{Open Research Questions and Future Steps} \label{future}  \label{questions}

This review has posed and answered some new key questions in the field. Nevertheless, there are still some questions which do not have a definite answer yet:

\begin{enumerate}

\item How to quantify the performance of a DT: Quantitative metrics are essential to understand the real impact of using a digital twin. It can also help in knowing how the performance of a digital twin can be improved. For a complete evaluation of the performance of DT and its suitability to the domain concerned, these error metrics have to be categorised as domain-dependent and  domain-independent  (such as \citep{Smarslok2013} proposed error metrics specific to an aircraft model).  This could also include uncertainty quantification to calculate the confidence levels of outputs of DT. Also, the self-evolving nature of DT can be implemented only if the DT can assess its own performance quantitatively.
	\item How to rectify data and model errors: Information shared across devices might contain noise and not always be accurate. Incorrect information models can be detrimental to the entire DT system. If the data or model fed to the DT is inaccurate, it can result in huge losses, due to incorrect predictions or test results (the human  in the end of the 'loop' plays an important role here in deciding whether to follow the DT's recommendation or not) To build resilient and self-evolving DT, more research is needed. Machine learning is a promising direction for this \citep{Cronrath2019}. These could include, anomaly detection (DT could check whether a particular data is misfed), and traceability (logs of the data used by DT to generate outputs could be available to check how DT generated an output. This could also help in improving the DT and for feature engineering in machine learning).  No error quantification metric: It is essential to quantify the error made by the DT in order to understand its performance. It also helps the managers in deciding how much to rely on the outcome of the DT, depending on its confidence levels. Without error quantification, to follow or not to follow the feedback from DT is taking a chance.

\end{enumerate}

Apart from tackling these issues, we believe the following future steps will enhance the advancement of DT
 and lead to its widespread useful implementation:
\begin{enumerate}
	\item A Formal Definition of DT: As a lack of consensus exists for the definition of digital twin, having a formal definition will help to clarify the concept and reach a universally accepted concept.
	\item IoT standards required for DT: Since DT relies heavily on IoT devices for capturing and sharing data, real-time synchronisation and monitoring, knowing what IoT standards are best-suited for these operations will enhance the acceptance of DT and make it easier for widespread adoption. \citet{lu2020digital} recognises the same need for a data communication standard.  
	\item Regulations at enterprise and global levels \citep{Tao}:  As many companies collaborate across industries, having proper legal-binding regulations on the data used in DT is crucial for smooth operation. This also applies to sites spread across the globe which need  to adhere to laws applicable of the particular country. 

	\item Liaising with domain experts to spread DT across sectors: Communicating with domain experts will facilitate easier implementation and acceptance of DT into new domains. As domain experts know all there is to know about domain-specific requirements, they can be very useful in designing the DT. Once a design is ready, the implementation of the DT framework can be managed by programmers and developers.
\end{enumerate}

\section{Conclusions} \label{concl}

There have been many papers defining the digital twin and its components, however, no universal model exists till date which is adopted widely. We reviewed the past work and defined our DT reference model, based on these past works which include important components. These components define the essence of a digital twin, such as the  bijective relation between DT and the physical asset. These components also highlight how to evaluate the performance and integrity of a DT system (testing and  security components). 

As a tool which can have many sub-components spread across collaborators and industry partners, developing regulations and security mechanisms is imperative for widespread of adoption of DT to overcome the concerns regarding  data sharing. Machine learning techniques like federated learning are promising in this direction too.

Digital twin is a powerful tool with great capabilities combining simulation, autonomy, agent-based modeling, machine learning, prototyping, optimisation and big data into one. Theses sub-systems can be tailored and prioritised depending on the application domain needs. However, the advancement and research in these subsystems at times create a hindrance for the development of DT.
We conclude that, as up-and-coming the DT technology is, there are certain technical  and domain-dependent challenges that still need to be addressed. The technology does depend on its counterparts of IoT, machine learning and data, however, a seamless integration of all these leads to the powerful and efficient product that a DT is.

\section*{Acknowledgements}
The work reported here was sponsored by Research England’s Connecting Capability Fund award CCF18-7157- Promoting the Internet of Things via Collaboration between HEIs and Industry (Pitch-In). We thank Kate Price Thomas for her suggestions for the literature review and industrial works, and Andy Gilchrist for the continued support of the project.



\bibliographystyle{elsarticle-harv} 
\bibliography{lib.bib}


\end{document}